\documentclass[10pt,twocolumn,letterpaper]{article}

\usepackage{cvpr}
\usepackage{times}
\usepackage{epsfig}
\usepackage{graphicx}
\usepackage{amsmath}
\usepackage{svg}
\usepackage{lipsum}
\usepackage{amssymb}
\newcommand{\smalltitle}[1]{\vspace{0.2em}\noindent \textbf{{#1}}}

\usepackage[pagebackref=true,breaklinks=true,letterpaper=true,colorlinks,bookmarks=false]{hyperref}
\usepackage[labelsep=period]{caption}
\newenvironment{myitemize}[1][]{
\begin{list}{{#1}} %
{
	\setlength{\leftmargin}{1.1em}
	\setlength{\topsep}{0em}
	\setlength{\itemsep}{-0.5em}
}}
{\end{list}}

\cvprfinalcopy 

\def\cvprPaperID{****} 

\ifcvprfinal\fi
\begin{document}

\title{Multi-Context Attention for Human Pose Estimation}
\author{
Xiao Chu$^{1}~\footnotemark[1]$ \quad Wei Yang$^{1}$
\thanks{ The first two authors contribute equally to this work.} \quad Wanli Ouyang$^{1}$ \quad Cheng Ma$^{2}$ \quad Alan L. Yuille$^{3}$ \quad Xiaogang Wang$^{1}$ \\
$^{1}$ The Chinese University of Hong Kong, Hong Kong SAR, China\\
$^{2}$~Tsinghua University, Beijing, China\\
$^{3}$~Johns Hopkins University, Baltimore, USA\\
$^{1}${\tt\small \{xchu, wyang, wlouyang, xgwang\}@ee.cuhk.edu.hk  } \\
$^{2}${\tt\small  macheng13@mails.tsinghua.edu.cn \quad $^{3}$ alan.yuille@jhu.edu}
}

\maketitle

\begin{abstract}
In this paper, we propose to incorporate convolutional neural networks with a multi-context attention mechanism into an end-to-end framework for human pose estimation. 
We adopt stacked hourglass networks to generate attention maps from features at multiple resolutions with various semantics. 
The Conditional Random Field (CRF) is utilized to model the correlations among neighboring regions in the attention map.
We further combine  the holistic attention model, which focuses on the global consistency of the full human body, and the body part attention model, which focuses on the detailed description for different body parts. Hence our model has the ability to focus on different granularity from local salient regions to global semantic-consistent spaces.
Additionally, we design novel Hourglass Residual Units (HRUs) to increase the receptive field of the network. These units are extensions of residual units with a side branch incorporating filters with larger receptive fields, hence features with various scales are learned and combined within the HRUs. 
The effectiveness of the proposed multi-context attention mechanism and the hourglass residual units is evaluated on two widely used human pose estimation benchmarks. Our approach outperforms all existing methods on both benchmarks over all the body parts. 
\end{abstract}


\vspace{-1em}
\section{Introduction}

\begin{figure}
	\begin{center}
	\includegraphics[width=1\linewidth]{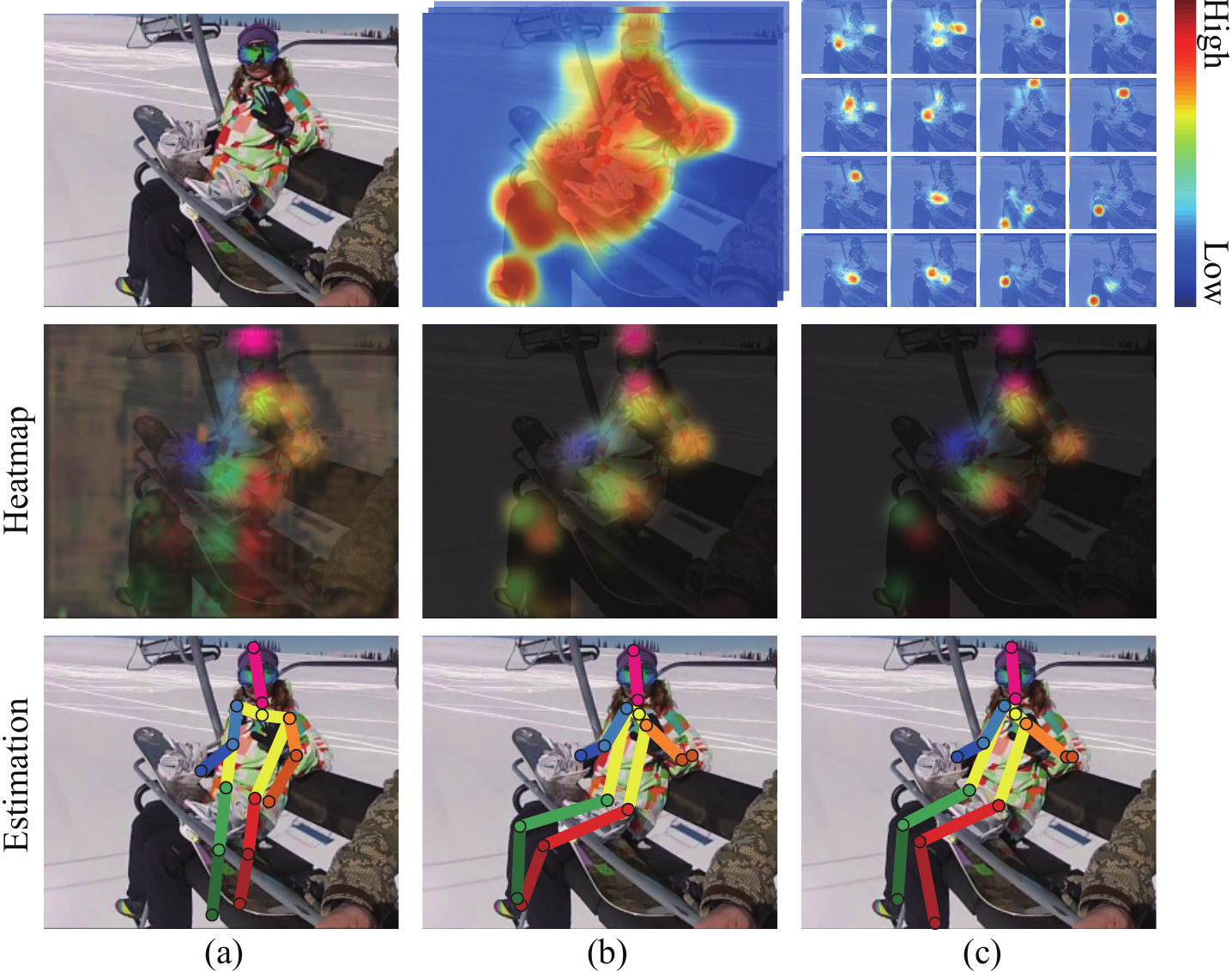}
	\end{center}
	\vspace{-1em}
	\caption{\small \textbf{Motivation}. The 1st row shows the input image, the holistic attention maps, and the part attention maps. The 2nd row shows the predicted heatmaps for part locations, where different colors correspond to different body parts. 
	The 3rd row visualizes the predicted poses. 
	We observe that (a) ConvNets may produce erroneous estimations due to cluttered background and self-occlusion. 
	(b) Visual attention provides an explicit way to model spatial relationships among human body parts, which is more robust. 
	(c) Part attention maps can help further refine the part locations by addressing the double counting problem. }
	\label{fig:motivation}
	\vspace{-1.5em}
\end{figure}

Human pose estimation is a challenging task in computer vision due to the articulation of body limbs, self occlusion, various clothing, and foreshortening. 
Significant improvements have been achieved by Convolutional Neural Networks (ConvNets)~\cite{tompson2014joint,toshev2014deeppose,chu2016structure,wei2016convolutional,tompson2015efficient,newell2016stacked}. 
However, for cluttered background with objects which are similar to body parts or limbs, or body parts with heavy occlusion, ConvNets may have difficulty to locate each body parts correctly, as demonstrated in Fig.~\ref{fig:motivation} (a). 
In the literature, the combination of multiple contextual information has been proved essential for vision tasks such as image classification~\cite{krizhevsky2012imagenet}, object  detection~\cite{girshick2014deformable,gidaris2015object,xyzeng2016object} and human pose estimation~\cite{ramakrishna2014pose, tompson2015efficient}. 
Intuitively, larger context region captures global spatial configurations of object, while smaller context region focuses on the local part appearance. 
However, previous works usually use manually designed multi-context representations, \eg, multiple bounding boxes~\cite{ramakrishna2014pose} or multiple image crops~\cite{krizhevsky2012imagenet}, and hence lack of flexibility and diversity for modeling the multi-context representations.

Visual attention is an essential mechanism of the human brain for understanding scenes effectively.
In this work, we propose to generate contextual representations with an attention scheme.
Instead of defining regions of interest manually by a set of rectangle bounding boxes, the attention maps are generated by an attention model, which depends on image features, and provide a principled way to focus on target regions with variable shapes. 
For example, an attention map focusing on the human body is shown in Fig.~\ref{fig:motivation} (b). 
It helps recover the missing body parts (\eg, \textit{legs}), and distinguishes the ambiguous background.
This allows the diversity of context to be increased, and so contextual region could be better adapted to each image. 
Furthermore, instead of adopting the spatial Softmax normalization widely used in conventional attention schemes, we design a novel attention model based on Conditional Random Fields, which is better in modeling the spatial correlations among neighboring regions.


The combination of multiple contextual information has been proved effective for various vision tasks~\cite{Zeng2013Multi,girshick2014deformable,ramakrishna2014pose,fan2015combining}.
To use the attention mechanism to guide multi-contextual representation learning, we adopt the stacked hourglass network structure~\cite{newell2016stacked}, which provides an ideal architecture to build a multi-context attention model. 
In each hourglass stack, features are pooled down to a very low resolution, then are upsampled and combined with high-resolution features. 
This structure is repeated for several times to gradually capture more global representations.
Within each hourglass stack, we first generate \textit{multi-resolution} attention maps from features of different resolutions.
Secondly, we generate attention maps for multiple hourglass stacks, which results in \textit{multi-semantics} attention maps with various levels of semantic meaning. 
Since these attention maps capture the configuration of the full human body, they are referred to as holistic attention models. 

%
While the holistic attention model is robust to occlusions and cluttered background, it lacks of precise description for different body parts. 
To overcome this limitation, we design a hierarchical visual attention scheme, which zooms in from holistic attention model to each body part, namely the \emph{part attention model}. 
This is helpful for precise localization of the body parts, as shown in Fig.~\ref{fig:motivation} (c).

Additionally, we introduce a novel ``Hourglass Residual Units" as a replacement for the residual unit~\cite{he2016deep} in our network.
It incorporates the expressive power of multi-scale features while preserving the benefit of residual learning. 
It also enables deep networks to have a faster growth of receptive field, which is essential for accurately locating body parts. 
When using these units within the ``macro" hourglass network, we obtain a \textit{nested} hourglass architecture. 
%


We show the effectiveness of the proposed end-to-end differentiable framework on two broadly used human pose estimation benchmarks, \ie, MPII Human Pose dataset~\cite{andriluka20142d} and the Leeds Sports Dataset~\cite{johnson2010clustered}. Our approach outperforms all the previous methods on both benchmarks for all the body parts.
The main contributions of this work are three folds:
\begin{myitemize}
\item[$\bullet$] 
We propose to use visual attention mechanism to automatically learn and infer the contextual representations, driving the model to focus on region of interest.
Instead of applying spatial Softmax normalization as in conventional attention models, we tailor the attention scheme for human pose estimation by introducing CRFs to model the spatial correlations among neighborhood joints.
To the best of our knowledge, this is the first attempt to utilize attention scheme for human pose estimation.

\item[$\bullet$] 
We use multi-context attention to make the model more robust and more accurate. Specifically, three types of attentions are designed, i.e., \textit{multi-resolution} attention within each hourglass, \textit{multi-semantics} attention across several stacks of hourglass, and a \textit{hierarchical} visual attention scheme to zoom in on local regions to see clearer.

\item[$\bullet$] 
We propose a generic hourglass residual unit (HRU), and build the \textit{nested hourglass networks} together with the stacked hourglass architecture. The HRUs incorporate features from different scales in the conventional residual unit. They also enable the network to see larger context in an earlier stage.

\end{myitemize}

\begin{figure*}[t]
\begin{center}
   \includegraphics[width=1\linewidth]{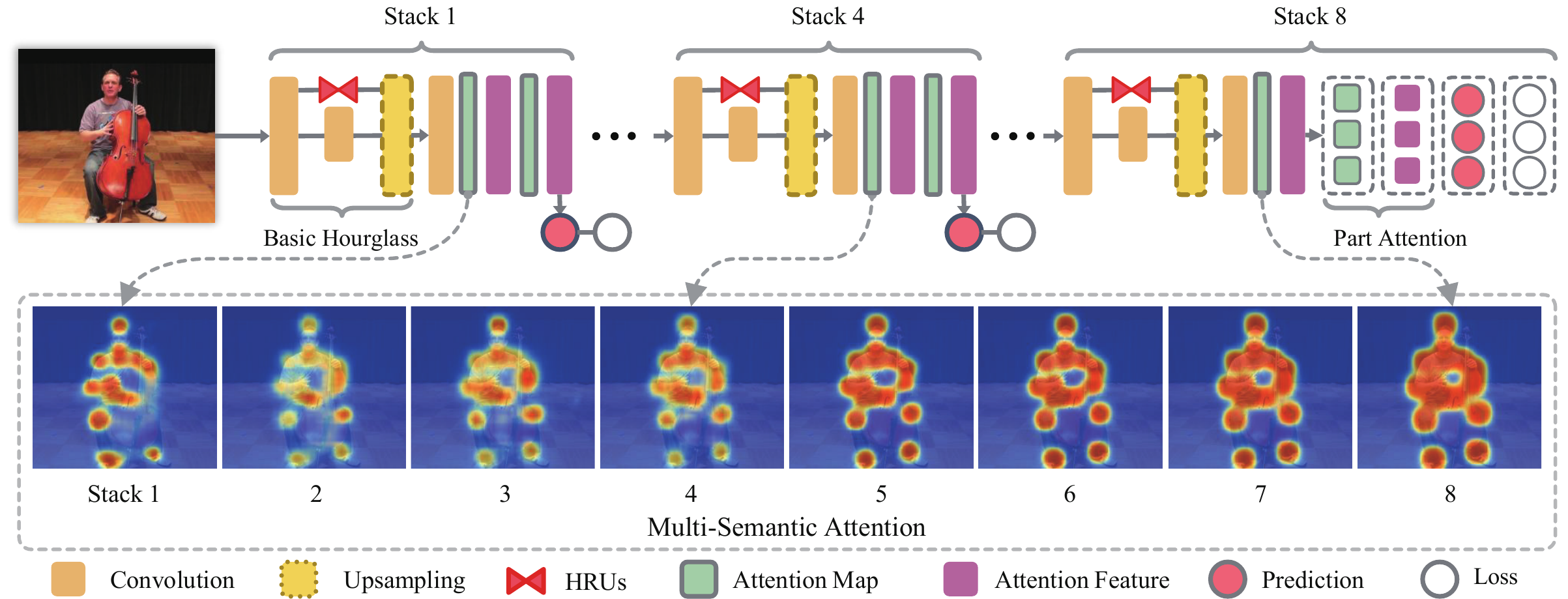}
\end{center}
	\vspace{-1em}
   \caption{\small \textbf{Framework}. 
   The basic structure is an 8-stack hourglass network. 
   In each stack of hourglass, we generate \textit{multi-resolution} attention maps. 
   We also apply \textit{multi-semantic} attention map to each hourglass as shown in stack 1 to stack 8.
   \textit{Hierarchical Attention Mechanism} for zooming in on local parts is applied in stack 5 to stack 8.}
   \vspace{-1.5em}
\label{fig:overview}
\end{figure*}

\section{Related Work}

\smalltitle{Human Pose Estimation} 
Articulated human poses were usually modeled by combination of unary term and graph models, \eg, mixture of body parts~\cite{yang2011articulated,chen2014articulated,ouyang2014multi} or pictorial structures~\cite{pishchulin2013poselet}. 
Recently, significant progresses have been achieved by introducing ConvNets for learning better feature representation~\cite{toshev2014deeppose,tompson2014joint,tompson2015efficient,chen2014articulated,wei2016convolutional,yang2016end,chu2016structure,chu2016crf,pishchulin2016deepcut,newell2016stacked}. 
For example, Chen and Yuille~\cite{chen2014articulated} introduced the ConvNet to learn both the unary and the pairwise term of a tree-structured graphical model. 
Tompson~\etal \cite{tompson2015efficient} used multiple branches of ConvNets to fuse the features from an image pyramid, and used a Markov Random Field (MRF) for post-processing. 
Convolutional Pose Machine~\cite{wei2016convolutional} incorporated the inference of the spatial correlations among body parts within the ConvNets.  State-of-the-art performance is achieved by the stacked hourglass network \cite{newell2016stacked} and its variant~\cite{bulat2016human}, which use repeated pooling down and upsampling process to learn the spatial distribution. 
Our approach is complementary to previous approaches by incorporating diverse image dependent multi-context representation to guide the human pose estimation.


\smalltitle{Multiple Contextual Information} 
The contextual information is generally referred to as regions surrounding the target locations~\cite{histogramsINRIA,fan2015combining,ramakrishna2014pose}, object-scene relationships~\cite{heitz2008learning, gupta2008context,divvala2009empirical}, and object-object interactions~\cite{yang2012recognizing}. It has been proved efficient in vision tasks as object classification~\cite{krizhevsky2012imagenet} and detection~\cite{Zeng2013Multi,histogramsINRIA,divvala2009empirical}. 
Recent works modeled contextual information by concatenating multi-scale features~\cite{girshick2014deformable,gidaris2015object}, or by gated functions to control the mutual influence of different contexts~\cite{xyzeng2016object}. 
The contextual regions, however, are manually defined as rectangles without considering the objects appearance. 
In this work, we adopt visual attention mechanism to focus on regions which are image dependent and adaptiving for multi-context modeling. Our approach increases the diversity of contexts.  
%


\smalltitle{Visual Attention Mechanism} 
Since the visual attention model is computationally efficient and is effective in understanding images, it has achieved great success in various tasks such as machine translation~\cite{bahdanau2014neural}, object recognition~\cite{ba2014multiple,gregor2015draw,cao2015look,xiao2015application}, image captioning~\cite{you2016image,xu2015show}, image question answering~\cite{yang2015stacked}, and saliency detection~\cite{kuen2016recurrent}. 
Existing approaches usually adopt recurrent neural networks to generate the attention map for an image region at each step, and combine information from different steps overtime to make the final decision~\cite{bahdanau2014neural, ba2014multiple,kuen2016recurrent}. 
To the best of our knowledge, our work is the first to investigate the use of attention models for human pose estimation. In addition, our design of the holistic attention map and the part attention map in learning attention in hierarchical order and the modeling of attention from different context and resolution are not investigated in these works.




\vspace{-1em}
\section{Framework}
\vspace{-0.5em}
An overview of our framework is illustrated in Fig.~\ref{fig:overview}. 
In this section, we briefly introduce the nested hourglass architecture, and the implementation of the multi-context attention model, including the multi-semantics, multi-resolution, and hierarchical holistic-part attention model. 
The generated attention maps are then used to reweight the features for automatically infer the regions of interest. 

\smalltitle{Baseline Network} 
We adopt an 8-stack hourglass network~\cite{newell2016stacked} as the baseline network. 
It allows for repeated bottom-up, top-down inference across scales with intermediate supervision at the end of each stack.  
In experiments, the input images are $256\times 256$, and the output heatmaps are $P\times 64\times 64$, where $K$ is the number of body parts. 
We follow previous work~\cite{tompson2015efficient,wei2016convolutional,newell2016stacked} to use the Mean Squared Error as the loss function. 

\smalltitle{Nested Hourglass Networks} 
We replace the residual units, which are along the side branches for combining features across multiple resolutions, 
by the proposed micro hourglass residual units (HRUs), and obtain a \textit{nested hourglass network}
, as illustrated in Fig.~\ref{fig: nested hourglass}. 
With this architecture, we enrich the information received by the output of each building block, which makes the whole framework more robust to scale change. 
Details of HRUs are described in Section~\ref{sec: Nested Hourglass unit}.

\smalltitle{Multi-Resolution Attention} 
Within each hourglass, the multi-resolution attention maps $\Phi_r$ are generated from features of different scales, where $r$ is the size of the features, as shown in Fig.~\ref{fig:multi-resolution}.
Attention maps are then combined to generate the refined features, which are further used to generate refined attention maps and further refined features, as shown in Fig.~\ref{fig:attention generation}.

\smalltitle{Multi-Semantics Attention}
Different stacks are with different semantics: lower stacks focus on local appearance, while higher stacks encode global representations.
Hence attention maps generated from different stacks also encode various semantic meanings. 
As shown in Fig.~\ref{fig:overview}, compare the \textit{left knee} in Stack 1 with 8, we can see that deeper stacks with global representations are able to recover occlusions.

\smalltitle{Hierarchical Attention Mechanism} 
In the lower stacks, \ie, stack 1 to stack 4, we use two holistic attention maps $\mathbf{h}_1^{\text{att}}$ and $\mathbf{h}_2^{\text{att}}$ to encode configurations of the whole human body. 
In the higher stacks, \ie, the 5th to the 8th stack, we design a hierarchical coarse-to-fine attention scheme to zoom into local parts. 

\vspace{-1em}
\section{Nested Hourglass Networks}\label{sec: Nested Hourglass unit}
In this section, we provide a detailed description of the proposed hourglass residual units (HRUs). We also provide comprehensive analysis of the receptive field.
\subsection{Hourglass Residual Units}

Let us first briefly recall Residual networks~\cite{he2016deep}.
Deep residual networks achieves compelling accuracy by an extremely deep stacks of ``Residual Units", which can be expressed as follows,
\begin{eqnarray}
\mathbf{x}_{n+1} = h(\mathbf{x}_n) + \mathcal{F}(\mathbf{x}_n, \mathbf{W}^{\mathcal{F}}_{n}), \label{eq:resU}
\end{eqnarray}
where $\mathbf{x}_n$ and $\mathbf{x}_{n+1}$ are the input and output of the $n$-th unit, 
and $\mathcal{F}$ is the stacked convolution, batch normalization, and ReLU nonlinearity. 
In~\cite{he2016deep}, $h(\mathbf{x}_n)=\mathbf{x}_n$ is the identity mapping.

\begin{figure}[t]
\begin{center}
   \includegraphics[width=0.9\linewidth]{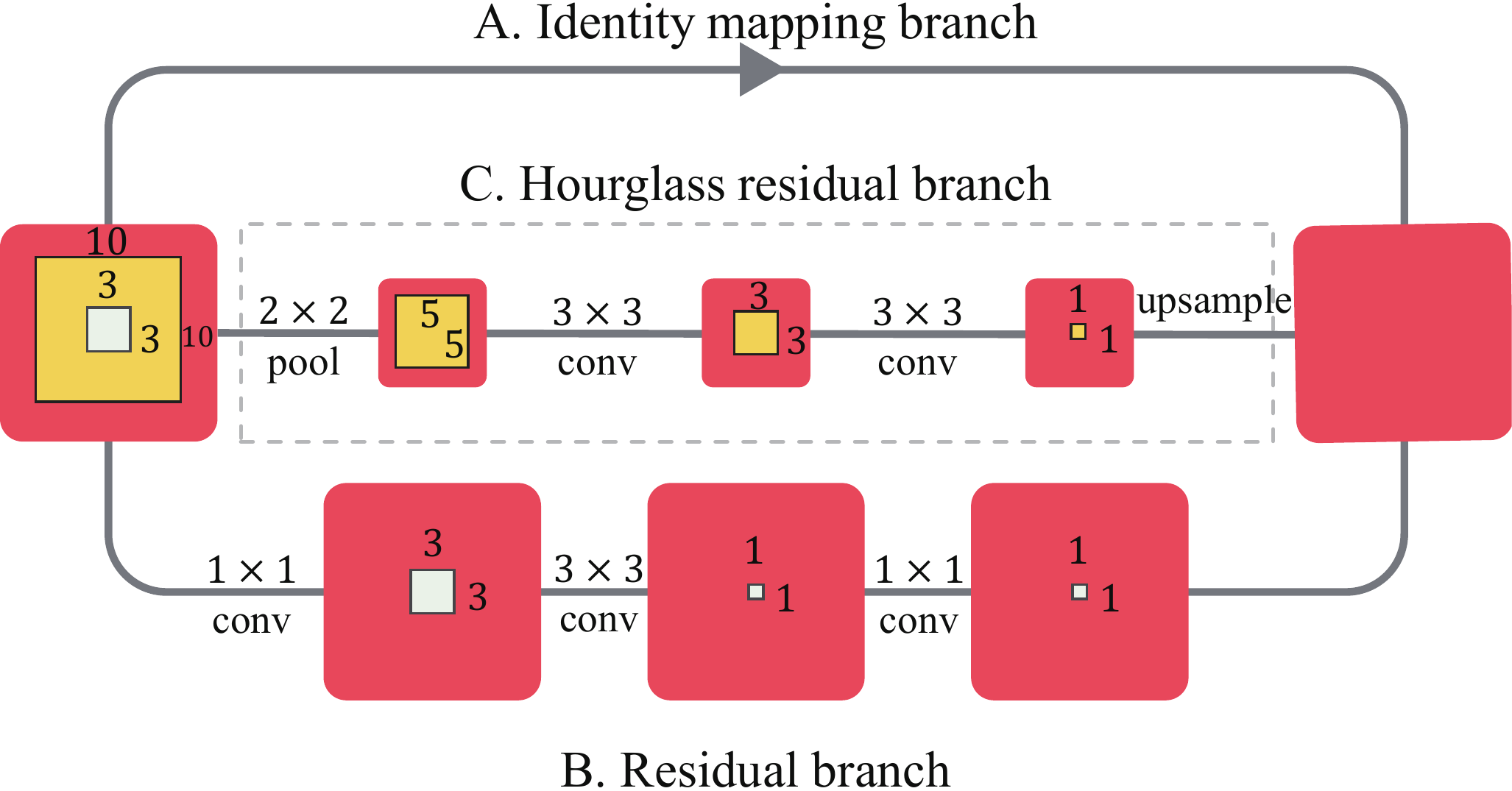}
\end{center}
  \vspace{-1.5em}
   \caption{\small An illustration of the \textit{hourglass residual unit}. It consists of three branches: (A) identity mapping, (B) residual branch, and (C) hourglass residual branch. The receptive field with respect to the input is $3\times 3$ and $10 \times 10$ for the conventional residual branch and the hourglass residual branch, respectively.}
\label{fig: nested hourglass}
  \vspace{-1em}
\end{figure}

In this paper, we focus on human pose estimation, which larger contextual regions are proved to be important for locating local body parts~\cite{wei2016convolutional,newell2016stacked}. 
The contextual region of a neuron is its corresponding receptive field. 
In this work, we propose to extend the original residual units by a micro hourglass branch. The resulted hourglass residual units (HRUs) have larger receptive field while preserve local details, as shown in Fig.~\ref{fig: nested hourglass}. 
We use this module in the stacked hourglass networks.
This architecture is referred to as ``nested hourglass networks'' because the hourglass structure is used at both the macro and micro levels. 
 
The mathematical formulation of our proposed HRUs is as follows:
\begin{eqnarray}
\mathbf{x}_{n+1} = \mathbf{x}_{n} + \mathcal{F}(\mathbf{x}_n, \mathbf{W}^{\mathcal{F}}_{n}) + \mathcal{P}(\mathbf{x}_{n}, \mathbf{W}^{\mathcal{P}}_{n}).
\label{eq:HRU}
\end{eqnarray}
Each HRU consists of three branches.
Branch (A), \ie $\mathbf{x}_{n}$ in (\ref{eq:HRU}), is the identity mapping. Hence, the property of ResNet in handling vanishing gradient is preserved in the HRUs.
Branch (B), \ie $\mathcal{F}(\mathbf{x}_n, \mathbf{W}^{\mathcal{F}}_{n})$ in (\ref{eq:HRU}),  is the residual block like the ResNet in (\ref{eq:resU}). 
Branch (C), \ie $\mathcal{P}(\mathbf{x}_{n}, \mathbf{W}^{\mathcal{P}}_{n})$ in (\ref{eq:HRU}), is our new design, which is a stack of a $2\times2$ max-pooling,  two $3\times 3$ convolutions followed by ReLU nonlinearity, and an upsampling operation.

\subsection{Analysis of Receptive Field of HRU}
The identity mapping in branch (A) has receptive size of one.
The residual block in branch (B) is a stack of convolutions ($\text{Conv}_{1\times1} + \text{Conv}_{3\times3} + \text{Conv}_{1\times1}$).
Hence, the neuron in the output feature corresponds to a $3 \times 3$ region of the input in this HRU. 
Branch (C) is our added branch. The structure of this branch is $\text{Pool}_{2\times 2} + \text{Conv}_{3\times3} + \text{Conv}_{3\times3} + \text{Deconv}_{2\times2}$. 
Due to max-pooling, the resolution for convolution in this branch is half of that in branches (A) and (B), and each neuron in the output feature map corresponds to a $10 \times 10$ region of the input, which is about 3 times the receptive field size of the residual block in branch (B).
These three branches, with different receptive fields and resolutions, are added together as the output of the HRU. 
Therefore, the HRU unit increases the receptive field size by including the branch (C) while preserves the high-resolution information by using branches (A) and (B). 

\section{Attention Mechanism}

We shall first briefly introduce the conventional soft attention mechanism, and then describe our proposed multi-context framework. 

\subsection{Conventional Attention}
Denote convolutional features by $\mathbf{f}$.
The first step in obtaining soft attention is to generate the summarized feature map as follows:
{\small \vspace{-0.8em}
\begin{equation}
   \mathbf{s} = g(\mathbf{W}^a * \mathbf{f}+\mathbf{b}),
   \label{eq:hidden map}
\end{equation}}
\!\!where $*$ denotes convolution, $\mathbf{W}^a$ denotes the convolution filters, and $g$ is the nonlinear activation function. $\mathbf{s}\in \mathbb{R}^{H\times W}$ summarizes information of all channels in $\mathbf{f}$.

Denote $\mathbf{s}(l)$ as the feature at location $l$ in the feature map $\mathbf{s}$, where $l=(x,y)$, $x$ is the horizontal location and $y$ is the vertical location. 
The Softmax operation is applied to $\mathbf{s}$ spatially as follows:
{\small \vspace{-0.8em}
\begin{equation}
   \Phi(l) = \frac{e^{\mathbf{s}(l)}}{\sum_{l'\in \mathbb{L}} e^{\mathbf{s}(l')}},
   \label{eq: spatial softmax}
\end{equation}}
\!\!where $\mathbb{L}=\{(x,y)| x=1, \ldots, W, y=1, \ldots, H\}$. $\Phi$ is the attention map, where $\sum_{l\in \mathbb{L}} \Phi(l)=1$.
Then the attention map is applied to the feature $\mathbf{f}$,
{\small \vspace{-1em}
\begin{equation}
\begin{split}
   \mathbf{h}^{\text{att}} &= {\Phi} \star \mathbf{f}, \qquad \textrm{where }  \mathbf{h}^{\text{att}}(c) =\mathbf{f}(c) \circ \Phi,
\end{split}
   \label{eq:AttFeat}
\end{equation} }
\!\!where $c$ is the index for feature channel. 
We use $\star$ to represent the channel-wise Hadamard matrix product operation. 
$\mathbf{h}^{\text{att}}$ is the refined feature map, which is the feature reweighted by the attention map, and has the same size as $\mathbf{f}$.


\subsection{Our Multi-Context Attention Model}
\begin{figure}[t]
\begin{center}
   \includegraphics[width=0.85\linewidth]{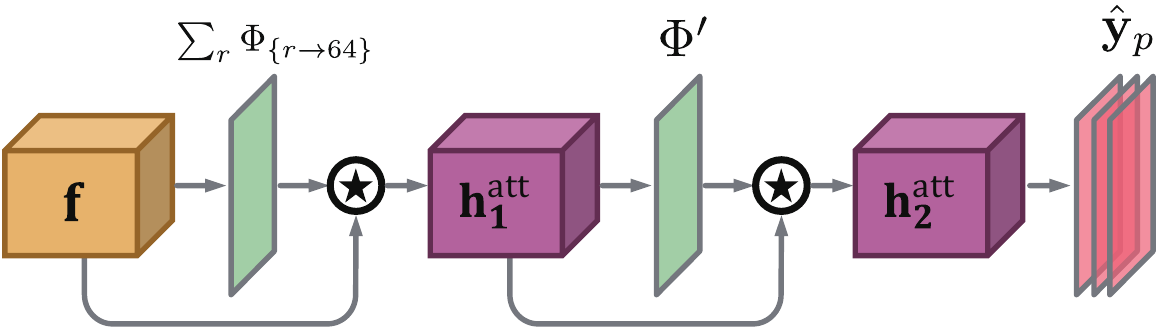}
\end{center}
	\vspace{-1.5em}
   \caption{\small An illustration of the attention scheme.}
   	\vspace{-1.5em}
\label{fig:attention generation}
\end{figure}

Our framework makes the following three modifications to the attention model. 
First, we replace the global Softmax in \ref{eq: spatial softmax} with a CRF to taking local pattern correlations into consideration.
Global spatial Softmax normalizes the whole image based on a constant factor, which ignores the local neighboring spatial correlations.
But we want attention maps to drive the network to concentrate on the complex human body configurations. 
More details are in Section \ref{Sec:Spatial}.
Second, we generate attention maps based on features of different resolutions to make the model more robust, as illustrated in Section \ref{Sec:Context}. 
Then multi-semantics attention is obtained by generating attention maps for each stack of the hourglass, as described in Section~\ref{Sec:multi-semantics}.
Finally, a hierarchical coarse to fine(\ie fully body to parts) attention scheme is used, to zoom into local part regions for more precise localization, which is introduced in Section \ref{Sec:partAtt}.
The whole framework is differentiable and trained end-to-end with random initialization. An illustration of our attention scheme is shown in Fig.~\ref{fig:attention generation}.
\vspace{-1em}
\subsubsection{Spatial CRF Model}
\label{Sec:Spatial}

In this work, we use Conditional Random Fields (CRFs) to model the spatial correlation. 
To make them differentiable, we use the mean-field approximation approach to recursively learn the spatial correlation kernel \cite{zheng2015conditional,krahenbuhl2012efficient}. 

The attention map is modeled as a two-class problem. Denote $y_l=\{0, 1\}$ as the attention label at the $i$-th location. In the CRF model, the energy of a label assignment $\mathbf{y}=\{y_l|l\in \mathbb{L}\}$ is as follows:
\begin{equation}\footnotesize
  E(\mathbf{z}) =  \sum_l y_l \psi_u(l) + \sum_{l,k} y_{l} w_{l,k} y_k,
\end{equation}
where $\psi(y_l) = g(\mathbf{h}, l)$ is the unary term that measures the inverse likelihood (and therefore, the cost) of the position
$l$ taking the attention label $y_l=1$. $w_{l,k}$ is the weight for compatibility between $y_l$ and $y_k$. Given the image $\mathbf{I}$, the probability of the label assignment $\mathbf{y}$  is  $P(\mathbf{y}|\mathbf{I}) = \frac{1}{Z}\exp(-E(\mathbf{y}|\mathbf{I}))$, where $Z$ is the partition function. 
The probability for $y_l=1$ is obtained iteratively using the mean-field approximation as follows:
\begin{equation}\small
  \Phi(y_l=1)_t = \sigma \left(  \psi_u(l) + \sum_{k} w_{l,k} \Phi(y_k=1)_{t-1} \right),
\end{equation}
where $\sigma(a) = 1/(1+\exp(-a))$ is the sigmoid function. 
$\psi_u(l)$ is obtained by convolution from features $\mathbf{h}$. $\sum_{k} w_{l,k} \Phi(y_j=1)$ is implemented by convolving the estimated attention map  $\Phi_{t-1}$ at the stage $t-1$ with the filters.  Initially, $\Phi(y_i=1)_1 =\sigma ( \psi_u(i))$.

In summary, the attention map $\Phi_t$ at the stage $t$ can be formulated as follows:
\begin{equation}\small
   \Phi_{t}= \mathcal{M}(\mathbf{s}, \mathbf{W}^{k}) = \left\{
   \begin{aligned}
      & \sigma (\mathbf{W}^{k} * \mathbf{s}) & t &= 0, \\
      & \sigma (\mathbf{W}^{k} * \Phi_{t-1}) & t & = 1,2,3,
   \end{aligned}
    \right.
    \label{eq:AttCNN}
\end{equation}
where $\mathcal{M}$ denotes a sequence of weights-sharing convolutions for the mean field approximation, 
$\mathbf{W}^k$ denotes the spatial correlation kernel. 
The $\mathbf{W}^k$ is shared across different time steps.
In our network, we use three steps of recursive convolution.


\vspace{-1em}
\subsubsection{Multi-Resolution Attention}
\label{Sec:Context}

\begin{figure} 
\begin{center}
   \includegraphics[width=1\linewidth]{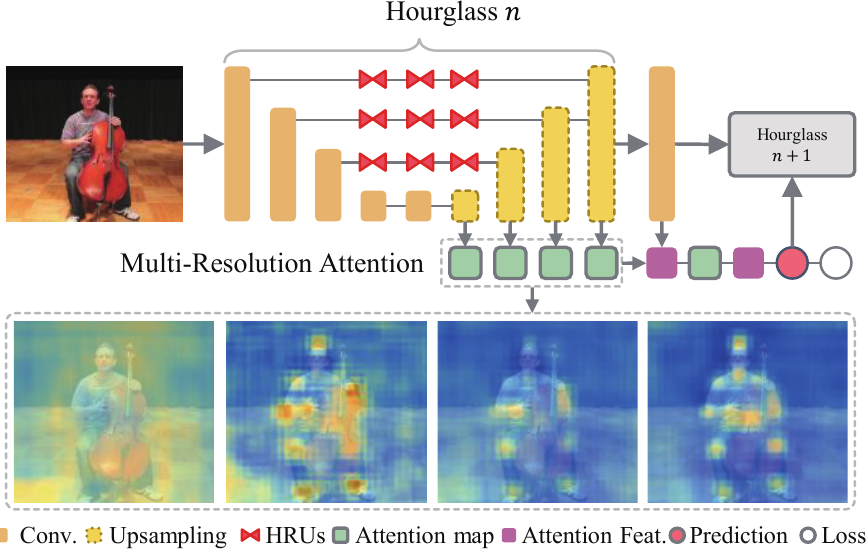}
\end{center}
	\vspace{-1.5em}
   \caption{\small The multi-resolution attention scheme within an hourglass. 
   In each stack of hourglass, we generate multi-resolution attention maps from features with different resolutions (a). 
   These maps are summed into a single attention map, which applies to features $\mathbf{f}$ to generate the refined feature $\mathbf{h}_1^{\text{att}}$.  
   }
   	\vspace{-1.5em}
\label{fig:multi-resolution}
\end{figure}

As shown in Fig.~\ref{fig:multi-resolution}, the up-sampling process generates features of different size r, \ie $\mathbf{f}_{r}$ for $r=8,16,32$ and 64. 
$\mathbf{s}_{r}$ is used to generate the attention map $\Phi_{r}$ using the procedure in (\ref{eq:AttCNN}). 
The attention map $\Phi_{r}$ is up-sampled to size 64, the up-sampled map is denoted by $\Phi_{\{r\rightarrow 64\}}$. 
These attention maps correspond to different resolutions. As shown in Fig.~\ref{fig:multi-resolution} (I), $\Phi_{\{8\rightarrow 64\}}$, which has lower resolution, and highlights the whole configure of human body.  
$\Phi_{64}$, which is generated with higher resolution, focusing on local body parts. 

 All up-sampled attention maps are summed up and then applied to the feature $\mathbf{f}$, 
\begin{equation}\small\vspace{-0.2em}
   \mathbf{h}_1^{\text{att}} = \mathbf{f} \star \left( \sum_{r=8,16,32,64} \Phi_{\{r \to64\}} \right),
   \label{eq:attCNN2}
\end{equation}
where the feature $\mathbf{f}$ is the output of the last layer in an hourglass stack as shown in Fig. \ref{fig:multi-resolution}. The operation $\star$ is illustrated in Eq.~(\ref{eq:AttFeat}).

The conventional way of using an attention map is to directly apply it to the feature which generates it. 
However, the features refined by attention map usually have large amount of values close to zero, and so a stack of many refined features makes the back-propagation difficult. 
To utilize information from multi-resolution features without sacrificing training efficiency, we generate attention maps from features with various resolutions, and apply them to the later features.

In addition to the multi-resolution attention, a refined attention map $\Phi'$ and its corresponding refined feature $\mathbf{h}_2^{\text{att}}$ are generated from $\mathbf{h}_1^{\text{att}}$, 
\begin{equation}
   \mathbf{h}_2^{\text{att}} = \mathbf{h}_1^{\text{att}} \star \Phi' = \mathbf{h}_1^{\text{att}} \star \mathcal{M}(\mathbf{h}_1^{\text{att}}, \mathbf{w}).
\end{equation}

\vspace{-1em}
\subsubsection{Multi-Semantics Attention} 
\label{Sec:multi-semantics}
The above procedure is repeated over stacks of hourglass to generate attention maps with multiple semantic meanings. 
Samples of $\Phi'$ are shown in Fig.~\ref{fig:overview} from stack 1 to 8. 
The attention maps at shallower hourglass stacks capture more local information.
For deeper hourglass stacks, the global information about the whole person is captured, which is more robust to occlusion. \

\subsubsection{Hierarchical Holistic-Part Attention}
\label{Sec:partAtt}

\begin{figure}[t]
\begin{center}
   \includegraphics[width=0.9\linewidth]{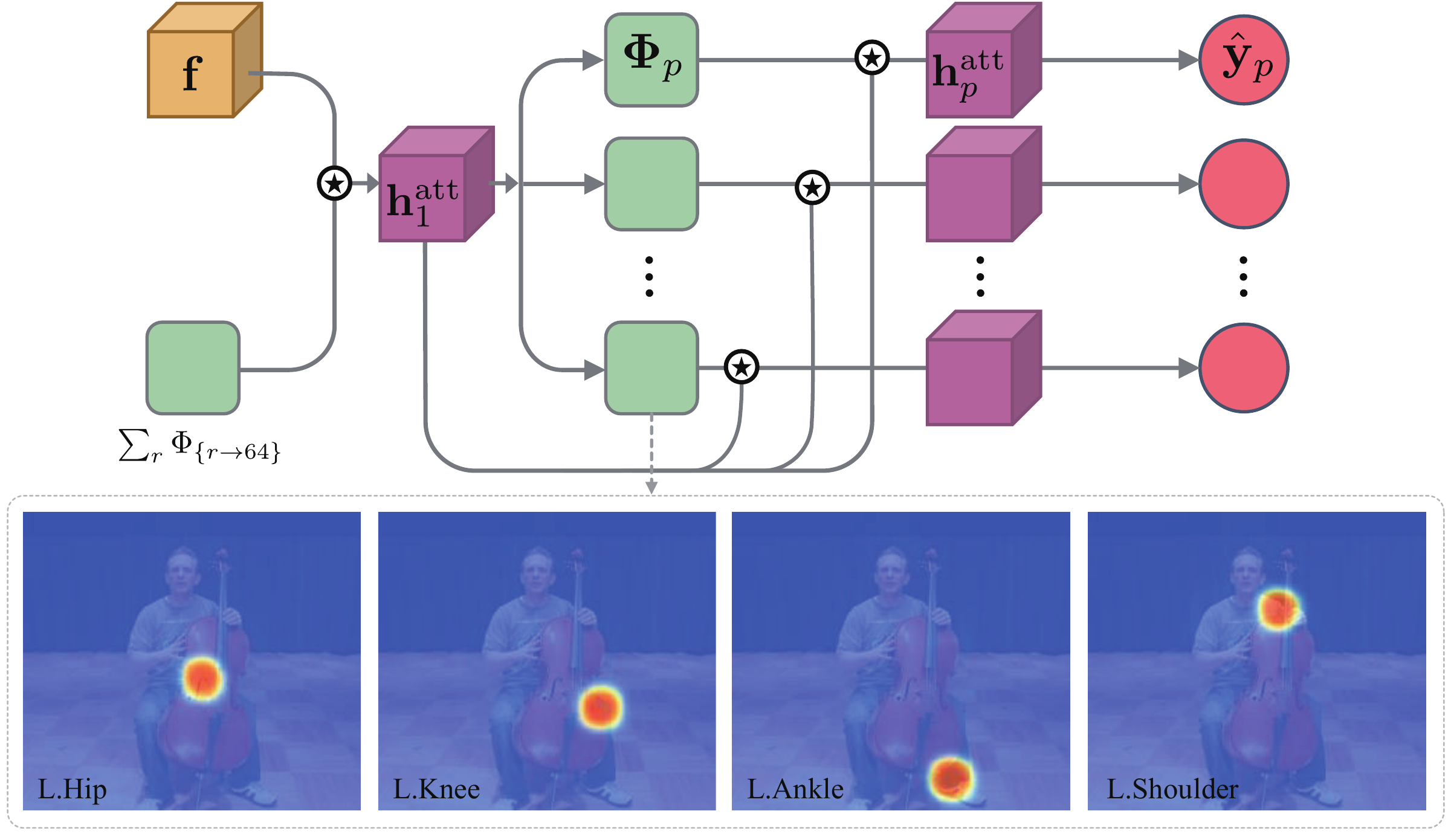}
\end{center}
	\vspace{-1.5em}
   \caption{\small Coarse-to-fine part attention model and the visualization of examplar part attention maps.}
   	\vspace{-1.5em}
\label{fig:part}
\end{figure}
 
In the 4th to 8th stacks of hourglass structure, we use the the refined feature $\mathbf{h}_1^{\text{att}}$ in Eq. (\ref{eq:attCNN2}) to generate the part attention maps as follows:
\begin{equation}\small
\begin{split}
   \mathbf{s}_p & =g(\mathbf{W}^{a}_p * \mathbf{h}^{\text{att}}_1+\mathbf{b}), \\
   \Phi_{p} & = \mathcal{M}(\mathbf{s}_p, \mathbf{W}^{k}_p),
  \label{eq:attPart}
\end{split}
\end{equation}
where $p \in\{1,\cdots, P\}$, 
$\mathbf{W}^a_{p}$ denotes the parameters for obtaining the summarization map $\mathbf{s}_p$ of part $p$,
$\mathbf{W}^k_{p}$ denotes the spatial correlation modeling for part $p$.
The part attention map $\Phi_{p}$ is combined with the refined  feature map $\mathbf{h}_1^{\text{att}}$ to obtain the refined  feature map for part $p$ as follows:
\begin{equation}\small
    \mathbf{h}_{p}^{\text{att}}=\mathbf{h}_1^{\text{att}} \star \Phi_{p}.
\end{equation}
The heatmap predication for the $p$th body joint is based on the refined  features $\mathbf{h}_{p}^{\text{att}}$,
\begin{equation}\small
   \mathbf{\hat{y}}_{p} = \mathbf{w}_p^{\text{cls}} * \mathbf{h}_{p}^{\text{att}},
\end{equation}
where $\mathbf{\hat{y}}_{p}$ is the heatmap for the $p$th part, $\mathbf{w}_p^{cls}$ is the classifier. 
In this way, we guarantee that the attention map $\Phi_{p}$ is specific for the body joint $p$. 
Some qualitative results of part attention maps are shown in Fig. \ref{fig:part}.

\section{Training the model}
Each stack in the hourglass produces the estimated heatmaps for the body joints.
We adopt the loss function in \cite{newell2016stacked} for learning the model. For each stack, the Mean Squared Error (MSE) loss is computed by
\begin{align}\small
L = \sum_{p=1}^{P} \sum_{l\in \mathbb{L}} \|\mathbf{\hat{y}}_{p} (l) - \mathbf{y}_{p} (l)\|_2^2
\end{align}
where $p$ denotes the $p$th body part, $l$ denotes the $l$th location.  $\mathbf{\hat{y}}_{p}$ denotes the predicted heatmap for part $p$, and $\mathbf{y}_{p}$ the corresponding ground-truth heatmap generated by a 2-D Gaussian centered on the body part location. 


The attention maps help to drive the network to focus on hard negative samples. 
After several stages of training, the attention maps fire on human body region, where the true positive samples are highlighted by attention maps. The refined  features are used for learning classifiers for the regions with human body, with easy background regions removed at the feature level by the learned attention maps. 
Consequentially, for part attention maps, the classifiers focus on classifying each body joint based on well defined human body regions, without considering the background.


\begin{figure}[t]
\begin{center}
  \includegraphics[width=1\linewidth]{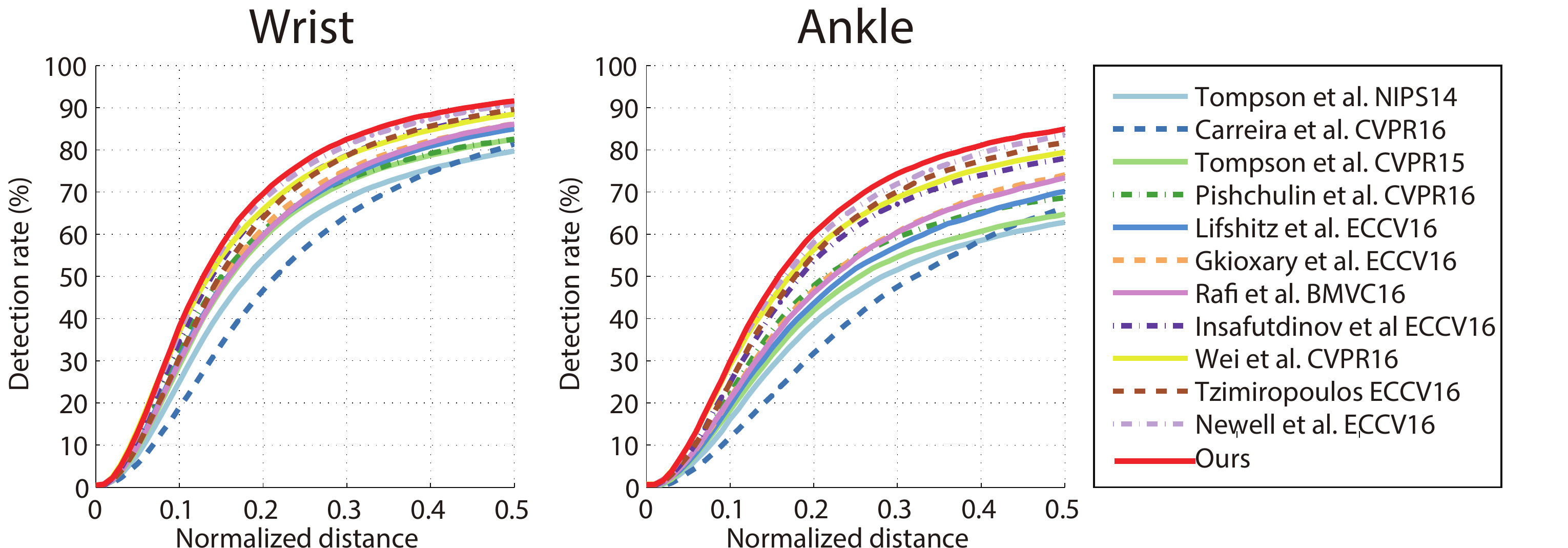}
\end{center}
\vspace{-1.2em}
   \caption{\small Comparisons of PCKh curve on the MPII Human Pose test set on the most challenging body joints, \ie, \textit{wrist} and \textit{ankle}. }
   \vspace{-1.1em}
\label{fig:comparisons_mpii}
\end{figure}

\section{Experiments}

\smalltitle{Dataset}
We evaluate the proposed method on two widely used benchmarks, MPII Human Pose~\cite{andriluka20142d} and extended Leeds Sports Poses  (LSP)~\cite{johnson2010clustered}. The MPII Human Pose dataset includes about 25k images with 40k annotated poses. The images were collected from YouTube videos covering daily human activities with highly articulated human poses. The LSP dataset consists of 11k training images and 1k testing images from sports activities. 

\smalltitle{Data Augmentation} During training, we crop the images with the target human centered at the images with roughly the same scale, and warp the image patch to the size $256\times 256$. Then we randomly rotate ($\pm 30^{\circ}$) and flip the images. We also perform random rescaling (0.75 to 1.25) and color jittering to make the model more robust to scale and illumination change. During testing, we follow the standard routine to crop image patches with the given rough position and the scale of the test human for MPII dataset. For the LSP dataset, we simply use the image size as the rough scale, and the image center as the rough position of the target human to crop the image patches. All the experimental results are produced from the original and flipped image pyramids with 6 scales.

\smalltitle{Experiment Settings} We train our model with Torch7~\cite{collobert2011torch7} using the initial learning rate of $2.5 \times 10^{-4}$. The parameters are optimized by RMSprop~\cite{tieleman2012lecture} algorithm. We train the model on the MPII dataset for 130 epochs and the LSP dataset for 60 epochs. We adopt the validation split for the MPII dataset used in~\cite{tompson2015efficient} to monitor the training process.

\subsection{Results}

\begin{table}
\begin{footnotesize}
  \centering
  \begin{tabular} 
  {@{}p{2.6cm}p{0.3cm}p{0.3cm}p{0.3cm}p{0.3cm}p{0.3cm}p{0.3cm}p{0.3cm}p{0.5cm}}
  \hline
  Method & Head & Sho. & Elb. & Wri. & Hip & Knee & Ank. & Mean\\
  \hline
  Pishchulin \etal~\cite{pishchulin2013strong} & 74.3  & 49.0  & 40.8  & 34.1  & 36.5  & 34.4 & 35.2 & 44.1  \\
  Tompson \etal~\cite{tompson2014joint}& 95.8  & 90.3  & 80.5  & 74.3  & 77.6  & 69.7 & 62.8 & 79.6  \\
  Carreira \etal~\cite{carreira2016human} & 95.7  & 91.7  & 81.7  & 72.4  & 82.8  & 73.2 & 66.4 & 81.3  \\
  Tompson \etal~\cite{tompson2015efficient}& 96.1  & 91.9  & 83.9  & 77.8  & 80.9  & 72.3 & 64.8 & 82.0  \\
  Hu\&Ramanan~\cite{hu2016bottom} & 95.0  & 91.6  & 83.0  & 76.6  & 81.9  & 74.5 & 69.5 & 82.4  \\
  Pishchulin \etal~\cite{pishchulin2016deepcut} & 94.1  & 90.2  & 83.4  & 77.3  & 82.6  & 75.7 & 68.6 & 82.4   \\
  Lifshitz \etal~\cite{lifshitz2016human} & 97.8  & 93.3  & 85.7  & 80.4  & 85.3  & 76.6 & 70.2 & 85.0   \\
  Gkioxary \etal~\cite{gkioxari2016chained} & 96.2  & 93.1  & 86.7  & 82.1  & 85.2  & 81.4 & 74.1 & 86.1   \\
  Rafi \etal~\cite{rafi2016efficient} & 97.2  & 93.9  & 86.4  & 81.3  & 86.8  & 80.6 & 73.4 & 86.3   \\
  Insafutdinov \etal~\cite{insafutdinov2016deepercut} & 96.8  & 95.2  & 89.3  & 84.4  & 88.4  & 83.4 & 78.0 & 88.5   \\
  Wei \etal~\cite{wei2016convolutional} & 97.8  & 95.0  & 88.7  & 84.0  & 88.4  & 82.8 & 79.4 & 88.5   \\
  Bulat\&Tzimiropoulos~\cite{bulat2016human} & 97.9  & 95.1  & 89.9  & 85.3  & 89.4  & 85.7 & 81.7 & 89.7   \\
  Newell \etal~\cite{newell2016stacked} & 98.2  & 96.3  & 91.2  & 87.1  & 90.1  & 87.4 & 83.6 & 90.9   \\
  \hline  
  Ours & \textbf{98.5}  & \textbf{96.3}  & \textbf{91.9}  & \textbf{88.1}  & \textbf{90.6}  & \textbf{88.0} & \textbf{85.0} & \textbf{91.5}  \\
  \hline
  \end{tabular}
  \vspace{-1em}
  \caption{\small Comparisons of PCKh@0.5 score on the MPII test set.}
  \label{tab:MPII}
  \vspace{-1em}
\end{footnotesize}
\end{table}

\begin{table} 
\begin{footnotesize}
  \centering
  \begin{tabular}{@{}p{2.7cm}p{0.3cm}p{0.3cm}p{0.3cm}p{0.3cm}p{0.3cm}p{0.3cm}p{0.3cm}p{0.4cm}}
  \hline
  Method & Head & Sho. & Elb. & Wri. & Hip & Knee & Ank. & Mean \\
  \hline 
  Belagiannis\&Zisserman~\cite{belagiannis2016recurrent} & 95.2 & 89.0 & 81.5 & 77.0 & 83.7 & 87.0 & 82.8 & 85.2 \\
  Lifshitz \etal~\cite{lifshitz2016human} & 96.8 & 89.0 & 82.7 & 79.1 & 90.9 & 86.0 & 82.5 & 86.7 \\
  Pishchulin \etal~\cite{pishchulin2016deepcut} & 97.0 & 91.0 & 83.8 & 78.1 & 91.0 & 86.7 & 82.0 & 87.1 \\
  Insafutdinov \etal~\cite{insafutdinov2016deepercut} & 97.4 & 92.7 & 87.5 & 84.4 & 91.5 & 89.9 & 87.2 & 90.1 \\
  Wei \etal~\cite{wei2016convolutional} & 97.8 & 92.5 & 87.0 & 83.9 & 91.5 & 90.8 & 89.9 & 90.5 \\
  Bulat\&Tzimiropoulos~\cite{bulat2016human} & 97.2 & 92.1 & 88.1 & 85.2 & 92.2 & 91.4 & 88.7 & 90.7 \\
  \hline
  Ours & \textbf{98.1} & \textbf{93.7} & \textbf{89.3} & \textbf{86.9} & \textbf{93.4} & \textbf{94.0} & \textbf{92.5} & \textbf{92.6} \\
  \hline      
  \end{tabular}
  \vspace{-1em}
  \caption{\small   Comparisons of PCK@0.2 score on the LSP dataset.}
  \label{tab:LSP}
  \vspace{-1.5em}
\end{footnotesize}
\end{table}

We use the Percentage Correct Keypoints (PCK)~\cite{yang2013articulated}  metric for comparisons on the LSP dataset, and the PCKh measure~\cite{andriluka20142d}, where the error tolerance is normalized with respect to head size, for comparisons on the MPII Human Pose dataset.

\smalltitle{MPII Human Pose}  
Table~\ref{tab:MPII} reports the comparison of the PCKh performance of our method and previous state-of-the-art at a normalized distance of $0.5$. Our method achieves state of the art $91.5\%$ PCKh scores. In particular, for the most challenging body parts, \eg, \textit{wrist} and \textit{ankle}, our method achieves $1.0\%$ and $1.4\%$ improvement compared with the closest competitor respectively, as shown in Fig.~\ref{fig:comparisons_mpii}.


\smalltitle{Leeds Sports Pose} 
We train our model by adding the MPII training set to the extended LSP training set with \textit{person-centric} annotations, which is a standard routine~\cite{wei2016convolutional,insafutdinov2016deepercut,pishchulin2016deepcut,lifshitz2016human, belagiannis2016recurrent}. Table~\ref{tab:LSP} reports the PCK at threshold of 0.2. Our approach outperforms the state-of-the-art across all the body joints, and obtains $1.9\%$ improvement in average.

\subsection{Component Analysis}

\begin{figure} 
  \begin{center}
    \includegraphics[width=1\linewidth]{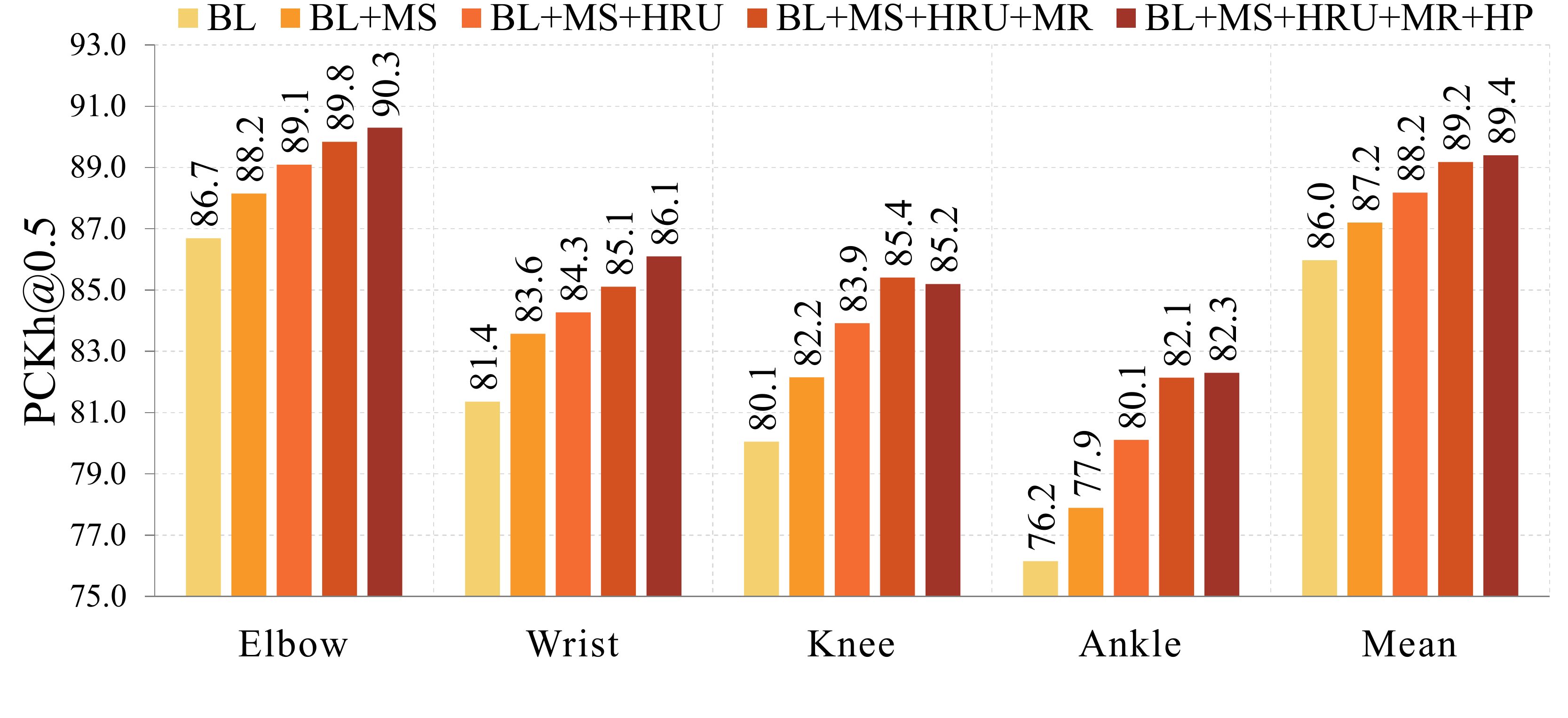}
  \end{center}
    \vspace{-2em}
  \caption{\small \textbf{Component analysis}. PCKh scores at threshold of 0.5 on the MPII validation set. }
    \vspace{-1.5 em}
  \label{fig:component_analysis}
\end{figure}

\begin{figure*} 
\begin{center}
  \includegraphics[width=1\linewidth]{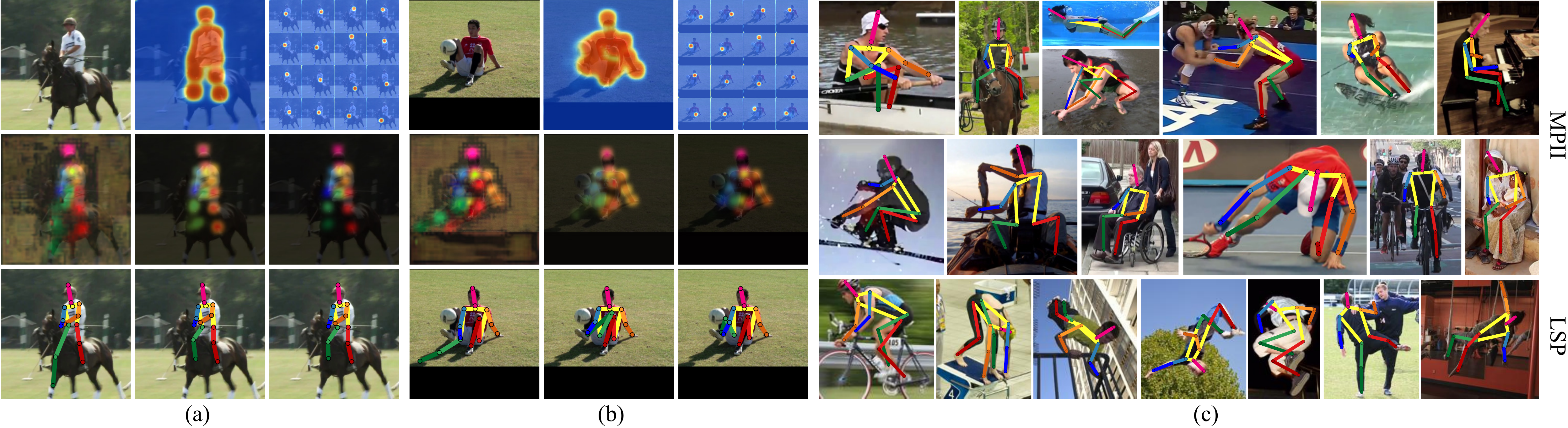}
\end{center}
\vspace{-1.5em}
   \caption{\small \textbf{Qualitative evaluation}. (a-b) 1st row to 3rd row: 2 input images, 4 attention maps, 6 heatmaps, and 6 predicted poses. (c) Examples of estimated poses on the MPII test set and the LSP test set (Best viewed in electronic form with $4\times$ zoom in).}
\vspace{-1em}
\label{fig:mpii_att_comparison}
\end{figure*}

To investigate the efficacy of the proposed multi-context attention mechanism and the hourglass residual unit, we conduct ablation experiments on the validation set~\cite{tompson2015efficient} of the MPII Human Pose dataset. We use an 8-stack hourglass network~\cite{newell2016stacked} as our baseline model if not specified. 
The overall result is shown in Fig.~\ref{fig:component_analysis}. 
Based on the baseline network (BL), we analyze each proposed component, \ie, the Multi-Semantics attention model (MS), Hourglass Residual Units (HRUs), Multi-Resolution attention model (MR), and the Hierarchical Part attention model (HP), by comparing the PCKh score. 

\smalltitle{Multi-Semantics Attention} 
We first evaluate the multi-semantics attention model. By adding holistic attention model at the end of each stack of hourglass (``BL+MS"), we get an $87.2\%$ PCKh score, which is a $1.2\%$ improvement compared to the baseline model. 

\smalltitle{Hourglass Residual Unit} 
To explore the effect of the residual pooling unit, we further use the HRUs to replace the original residual units when combining features from different resolutions (``BL+MS+HRU"), as illustrated in Fig.~\ref{fig:overview}. The addition of hourglass residual unit result in a further $1\%$ improvement. 
As discussed in~\cite{newell2016stacked}, improvements cannot be easily obtained by simply stacking more than eight hourglass modules. 
We provide a way to increase the model capacity effectively.

\smalltitle{Multi-Resolution Attention} 
By generating attention maps from features with multiple resolution (``BL+MS+HRU+MR"), our method obtains a further $1\% $ improvements.   

\smalltitle{Hierarchical Attention} 
We also show the improvement brought by the hierarchical holistic-local attention model. We replace the refined holistic attention map by a set of part attention maps from stack four to eight, and obtain the highest mean PCKh score $89.4\%$. 
We observe the improvements are mostly brought by the refined localization of body parts. 
In some cases, the part attention model could even correct the double counting problem, as demonstrated in Fig.~\ref{fig:motivation} (c).

\begin{figure} 
  \begin{center}
    \includegraphics[width=1\linewidth]{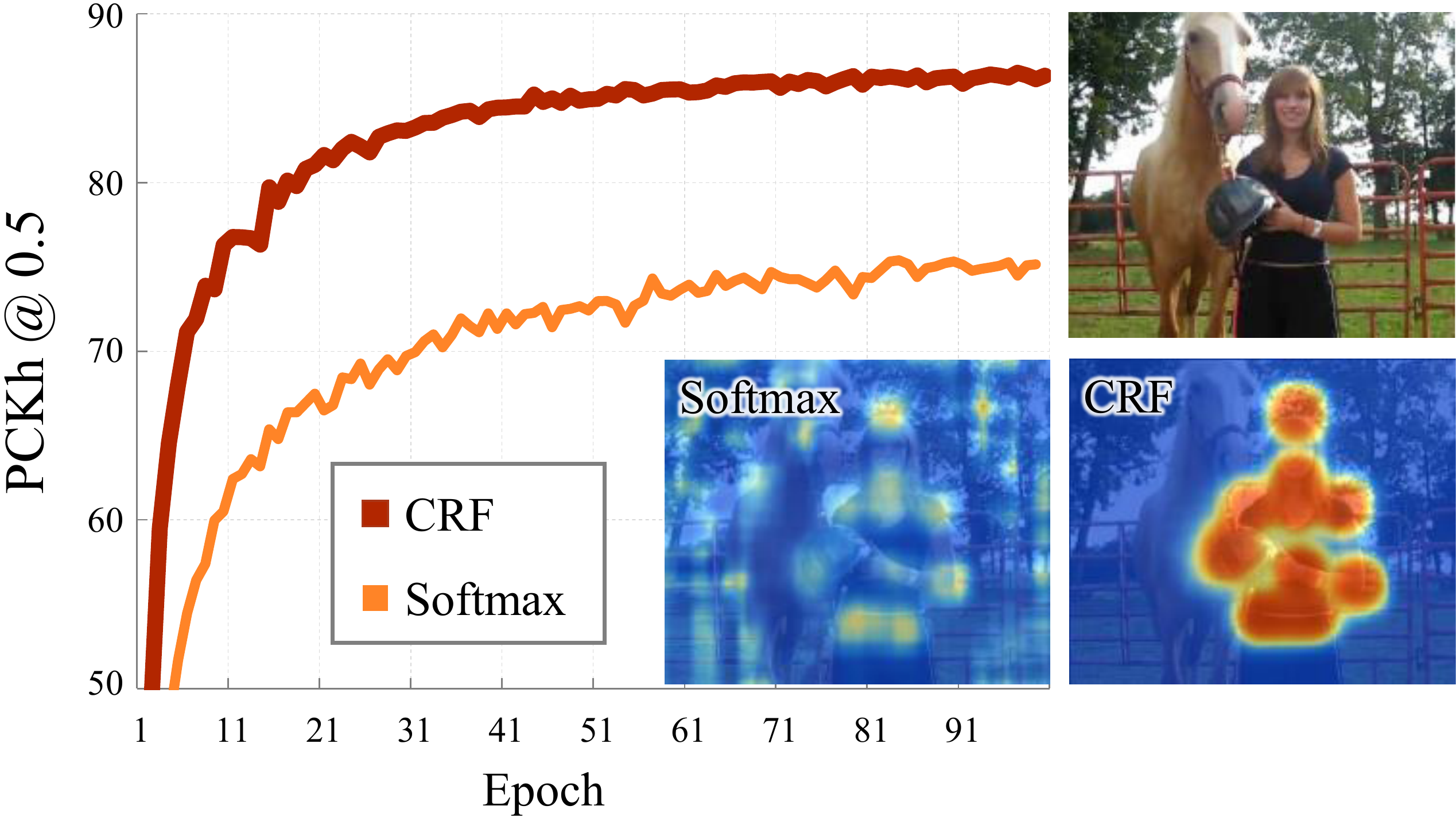}
  \end{center}
  \vspace{-2em}
  \caption{\small PCKh@0.5 on the MPII validation set across training.}
    \vspace{-1.5em}
  \label{fig:valid_acc}
\end{figure}

\smalltitle{Softmax vs. CRF} Finally, we compare the proposed CRF spatial attention model with the conventional Softmax attention model based on a 2-stack hourglass network. We compare the accuracy rates, \ie, PCKh at 0.5, on the validation set as training progresses in Fig.~\ref{fig:valid_acc}. The CRF attention model converges much faster and achieves higher validation accuracy than the Softmax attention model. 
We visualize the attention maps generated by these two models, and observe that CRF attention models generates much more cleaner attention maps compared with Softmax attention model due to its better ability to model spatial correlations among body parts.

\subsection{Qualitative Results}
To gain insights on how attention works, we compare the baseline model with the proposed model by visualizing the attention maps, the score maps, and the estimated poses, as demonstrated in Fig.~\ref{fig:mpii_att_comparison} (a-b). 
We observe the baseline model may has difficulty in distinguishing objects with similar appearance with limbs (\eg, the horse leg in Fig.~\ref{fig:mpii_att_comparison} (a)), 
and the heavy shadow with ambiguous shape (Fig.~\ref{fig:mpii_att_comparison} (b)). 
So the holistic attention maps would be great help for removing cluttered background and reducing ambiguity. For part attention maps, besides providing more precise localization for the body parts, they could even help reduce the double counting problem. For example, the left and right \textit{ankle} can be distinguished by incorporating the part attention maps.

\begin{figure}
  	\begin{center}
	\includegraphics[width=1\linewidth]{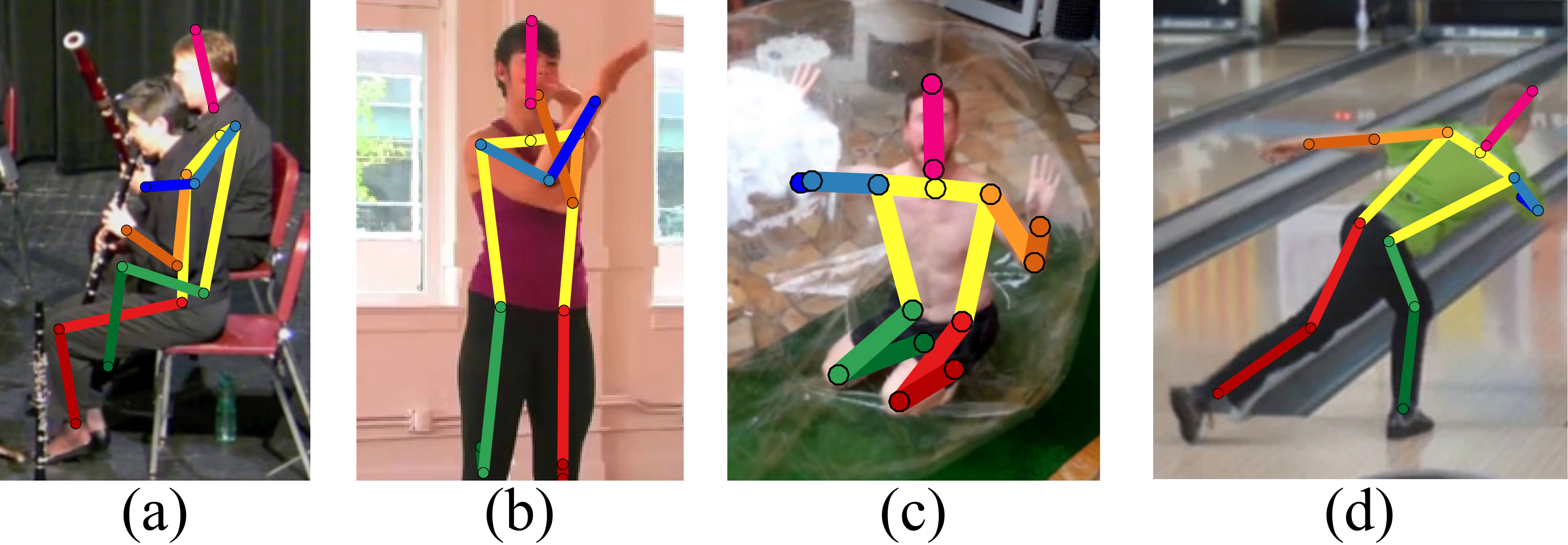}
  	\end{center}
  	\vspace{-2em}
    \caption{\small Failure cases caused by (a) overlapping people,  (b) twisted limbs,  (c) illumination, and  (d) left/right confusion.}
      	\vspace{-1.5em}
  	\label{fig:failure_cases}
\end{figure}

Fig.~\ref{fig:mpii_att_comparison} (c) demonstrates the poses predicted by our methods on the MPII test set and the LSP test set. Our method is robust to extremely difficult cases, \eg, rare poses, cluttered background, and foreshortening. 
However, as shown in Fig.~\ref{fig:failure_cases}, our method may fail in some cases which are also difficult for human eyes, \ie (a) heavy occlusion and ambiguity,  (b) twisted limbs,  (c) significant illumination change, and  (d) left/right body confusion caused by clothing/lighting. Please refer to the supplementary materials for more visualized results.



\vspace{-0.5em}
\section{Conclusion}
This paper has proposed incorporating multi-context attention and ConvNets into an end-to-end framework. 
We use visual attention to guide context modeling. 
Hence our framework has large diversity in contextual regions. 
Instead of using global Softmax, we introduce CRF for spatial correlation modeling. 
We build multi-context attention model along three components, 
\ie, multi-resolution, multi-semantics, and  hierarchical holistic-part attention scheme.
%
Additionally, an hourglass residual unit was proposed to enrich the expressive power of conventional residual unit. 
The proposed multi-context attention and the HRUs are general, and would help other vision tasks.
{\small
\bibliographystyle{ieee}
\bibliography{egbib}
}

\end{document}